\documentclass{article}

\PassOptionsToPackage{numbers}{natbib} 

\usepackage[preprint]{neurips_2021}

\usepackage[utf8]{inputenc} 
\usepackage[T1]{fontenc}    
\usepackage{hyperref}       
\usepackage{url}            
\usepackage{booktabs}       
\usepackage{amsfonts}       
\usepackage{nicefrac}       
\usepackage{microtype}      
\usepackage{xcolor}         
\usepackage{graphicx}
\usepackage{amsmath}
\usepackage{bbm}
\usepackage{bm}	

\usepackage[ruled,vlined]{algorithm2e}
\SetKwInOut{Input}{$\triangleright$\ Input}
\SetKwInOut{Parameters}{$\triangleright$\ Parameters}
\SetKwInOut{Output}{$\triangleright$\ Output}
\SetKwComment{Comment}{$\triangleright$\ }{}

\SetCommentSty{comsty}

\let\hat\widehat

\title{Soft-SVM Regression For Binary Classification}

\author{
  Man Huang \\
  Department of Mathematics \& Statistics\\
  Boston University\\
  Boston, MA 02215 \\
  \texttt{manhuang@bu.edu} \\
   examples of more authors
  \And
  Luis Carvalho \\
  Department of Mathematics \& Statistics\\
  Boston University\\
  Boston, MA 02215 \\
  \texttt{lecarval@math.bu.edu} \\
}

\begin{document}
\maketitle
\begin{abstract}
The binomial deviance and the SVM hinge loss functions are two of the most
widely used loss functions in machine learning. While there are many
similarities between them, they also have their own strengths when dealing
with different types of data. In this work, we introduce a new exponential
family based on a convex relaxation of the hinge loss function using
softness and class-separation parameters. This new family, denoted Soft-SVM,
allows us to prescribe a generalized linear model that effectively bridges
between logistic regression and SVM classification. This new model is
interpretable and avoids data separability issues, attaining good fitting and
predictive performance by automatically adjusting for data label separability
via the softness parameter. These results are confirmed empirically through
simulations and case studies as we compare regularized logistic, SVM, and
Soft-SVM regressions and conclude that the proposed model performs well in
terms of both classification and prediction errors.
\end{abstract}

\section{Introduction}
Binary classification has a long history in supervised learning, arising in
multiple applications to practically all domains of science and motivating the
development of many methods, including logistic regression, $k$-nearest
neighbors, decision trees, and support vector machines~\cite{hastie09}.
Logistic regression, in particular, provides a useful statistical framework
for this class of problems, prescribing a parametric formulation in terms of
features and yielding interpretable results~\cite{mccullagh89}. It suffers,
however, from ``complete separation'' data issues, that is, when the classes
are separated by a hyperplane in feature space. Support vector machines (SVMs),
in contrast, aim at finding such a separating hyperplane that maximizes its
margin to a subset of observations, the support
vectors~\cite{vapnik13,moguerza06}.
While SVMs are robust to complete separation issues, they are not directly
interpretable~\cite{lin02}; for instance, logistic regression provides class
probabilities, but SVMs rely on post-processing to compute them, often using
Platt scaling, that is, adopting a logistic transformation.

Both SVM and logistic regression, as usual in statistical learning methods,
can be expressed as optimization problems with a data fitting loss function
and a model complexity penalty loss. In Section~\ref{sec:methods} we adopt
this formulation to establish a new loss function, controlled by a
convex relaxation ``softness'' parameter, that comprehends both methods.
Adopting then a generalized linear model formulation, we further explore this
new loss to propose a new exponential family, \emph{Soft-SVM}. Given that this
new regularized regression setup features two extra parameters, we expect it to
be more flexible and robust, in effect bridging the performance from SVM and
logistic regressions and addressing their drawbacks. We empirically validate
this assessment in Section~\ref{sec:experiments} with a simulation study and
case studies. We show, in particular, that the new method performs well,
comparably to the best of SVM and logistic regressions and with a similar
smaller computational cost of logistic regression.

\section{The Soft-SVM Family}
\label{sec:methods}
\begin{figure}
\centering
\includegraphics[width=0.9\textwidth]{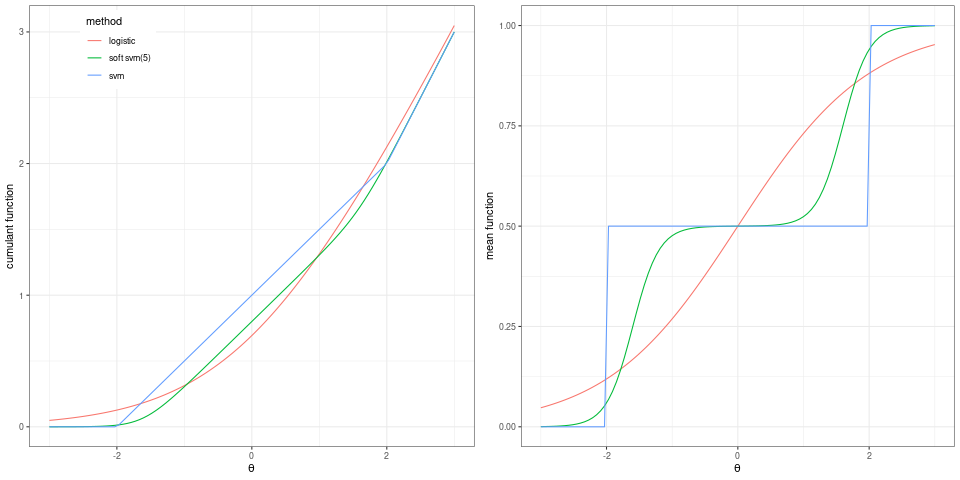}
\caption{Cumulant (left) and mean (right) functions for logistic, SVM, and Soft-SVM(5, 0.8).}
\label{fig:families}
\end{figure}
Suppose we have $n$ observations with $f$ features $x_i$ and labels
$y_i \in \{0, 1\}$, for $i = 1, \ldots, n$. In logistic regression we aim to
find a set of coefficients $\beta_0$ and $\beta$ that maximizes, with
$\theta_i \doteq \beta_0 + x_i^\top \beta$,
\begin{equation}
\label{eq:logistic-lhood}
\ell(\theta; \mathbf{y}) = \sum_{i = 1}^n y_i \theta_i - b(\theta_i),
\end{equation}
the log-likelihood (up to a constant). Parameters $\theta_i$ are canonical for
the Bernoulli family and thus coincide with the linear predictor
$\eta_i = \beta_0 + x_i^\top \beta$; function $b$ is the Bernoulli cumulant
function, given by $b(\theta) = \log(1 + e^\theta)$.
Similarly, in SVM regression, the goal is to maximize, again with respect to
$\beta_0$ and $\beta$, the negative cost
\begin{equation}
\label{eq:svm-lhood}
\sum_{i = 1}^n -[1 - (2 y_i - 1) \eta_i]_+ - \frac{\lambda}{2} \|\beta\|_2^2
=
\sum_{i = 1}^n y_i[(1 + \eta_i)_+ - (1 - \eta_i)_+] - (1 + \eta_i)_+
- \frac{\lambda}{2} \|\beta\|_2^2,
\end{equation}
where $x_+ = \max\{0, x\}$ is the plus function and $\lambda$ is a penalty
parameter. The data fitting loss per observation is $[1 - (2y_i - 1)\eta_i]+$,
the hinge loss. The quadratic complexity penalty here arises from a Lagrange
multiplier version of the original max-margin support vector
formulation~\cite{hastie09}. The negative loss in~\eqref{eq:svm-lhood} can
then be similarly expressed as a log-likelihood in~\eqref{eq:logistic-lhood}
by setting the canonical parameters and SVM cumulant function $s$ as
\[
\theta_i = (1 + \eta_i)_+ - (1 - \eta_i)_+
\quad \text{and} \quad
s(\theta) = (1 + \eta)_+ = \frac{1}{2}[(\theta + 2)_+ + (\theta - 2)_+].
\]
This, however, does not correspond to a valid exponential family specification
since both functions are not differentiable.
To address this issue and define a new exponential family, let us focus on the
cumulant function. We first adopt the convex relaxation
$p_\kappa(x) = \log(1 + \exp\{\kappa x\}) / \kappa$ for the plus function, the
soft-plus with softness parameter $\kappa$. Note that $p_\kappa(x) \rightarrow
x_+$ as $\kappa \rightarrow \infty$. Motivated by the property of both
logistic and SVM cumulants that $b(\theta) = \theta + b(-\theta)$, we define
the Soft-SVM canonical parameters and cumulant as
\begin{equation}
\label{eq:softsvm-cumulant}
\theta_i = p_\kappa(\eta_i + \delta) - p_\kappa(\eta_i - \delta)\doteq f_{\kappa,\delta}(\eta_i)
\quad \text{and} \quad
b_{\kappa,\delta}(\theta) = \frac{1}{2}[p_\kappa(\theta + 2\delta) +
p_\kappa(\theta - 2\delta)],
\end{equation}
where $\delta$ is a separation parameter in the cumulant. Parameter $\delta$
works together with the softness parameter $\kappa$ to effective bridge
between logistic and SVM regressions: when $\kappa=1$ and $\delta = 0$, we
have logistic regression, while as $\kappa \rightarrow \infty$ and $\delta
\rightarrow 1$ we obtain SVM regression.
This way, $b_{1,0} \equiv b$ and $b_{\kappa,\delta} \rightarrow s$ pointwise as
$\kappa \rightarrow \infty$ and $\delta \rightarrow 1$.
The Soft-SVM penalized log-likelihood is then
\[
\ell_{\kappa,\delta}(\theta; \mathbf{y}) \doteq
\sum_{i=1}^n y_i \theta_i - b_{\kappa,\delta}(\theta_i) -
\frac{\lambda}{2}\|\beta\|_2^2 =
\sum_{i=1}^n y_i \theta_i - b_{\kappa,\delta}(\theta_i) -
\frac{\lambda}{2} \begin{bmatrix}\beta_0 \\ \beta\end{bmatrix}^\top
\begin{bmatrix} 0 & \\ & I_f\end{bmatrix}
\begin{bmatrix}\beta_0 \\ \beta\end{bmatrix},
\]
where we make clear that the bias $\beta_0$ is not penalized when using the
precision matrix $0 \oplus I_f$.

While the separation parameter $\delta$ has a clear interpretation from the
cumulant function, we can also adopt an equivalent alternative
parameterization in terms of a \emph{scaled} separation parameter $\alpha
\doteq \kappa \delta$ since the shifted soft plus can be written as
\[
p_\kappa(x + z\delta) =
\frac{1}{\kappa}\log\big(1 + e^{\kappa(x + z\delta)}\big) =
\frac{1}{\kappa}\log\big(1 + e^{\kappa x + z\alpha}\big).
\]
This alternative parameterization is more useful when we discuss the variance
function in Section~\ref{sec:varfun}.

\begin{figure}
\centering
\includegraphics[width=0.9\textwidth]{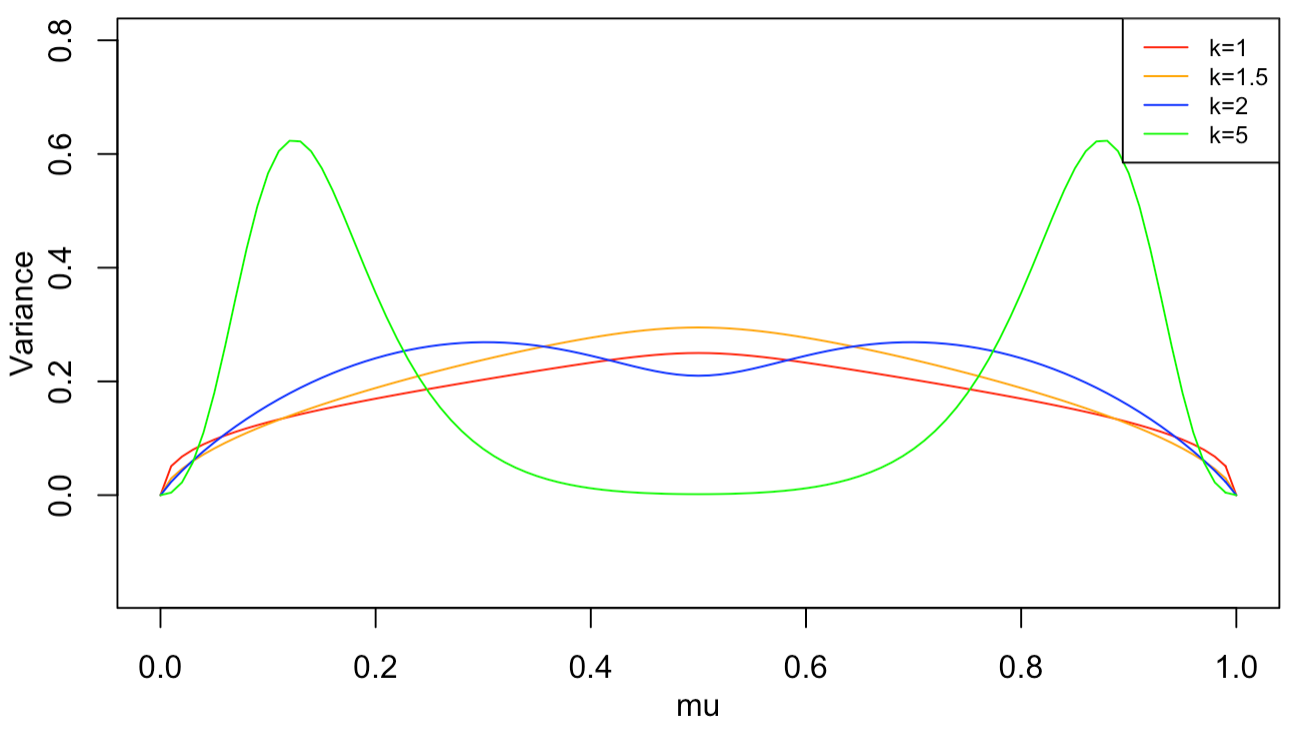}
\caption{Plot of variance functions when $\kappa \in \{1, 1.5, 2, 5\}$ and
$\delta = 1 - 1 / \kappa = \{0, 1/3, 1/2, 4/5\}$ respectively.}
\label{fig:variance}
\end{figure}

The mean function corresponding to $b_{\kappa,\delta}$ is its first derivative,
\begin{equation}
\label{eq:softsvm-mean}
\mu(\theta) = b_{\kappa,\delta}'(\theta) = \frac{1}{2}[
\text{expit}\{\kappa(\theta + 2\delta)\} +
\text{expit}\{\kappa(\theta - 2\delta)\}].
\end{equation}
With $v(x) = \text{expit}(x) (1 - \text{expit}(x)) = e^x / (1 + e^x) ^ 2$, the
variance function is
\begin{equation}
\label{eq:softsvm-var}
V_{\kappa,\delta}(\mu) = b_{\kappa,\delta}''\big([b^{'}_{\kappa,\delta}(\mu)]^{-1}) =
\frac{\kappa}{2}\big[
v(\kappa ([b^{'}_{\kappa,\delta}(\mu)]^{-1} + 2\delta)) +
v(\kappa ([b^{'}_{\kappa,\delta}(\mu)]^{-1} - 2\delta))
\big],
\end{equation}
where the inverse mean function $[b^{'}_{\kappa, \delta}(\mu)]^{-1}$ is given in
the next section.
Figure~\ref{fig:variance} shows the plot of variance function when $\kappa$
equals to different values and assumes the corresponding value of $\delta$ is
a function of $\kappa$: $\delta=1-1/\kappa$.

Finally, to relate the canonical parameters $\theta_i$ to the features, our link function $g$ can be found by solving the composite function $f^{-1}_{\kappa,\delta}\circ[b^{'}_{\kappa,\delta}(\mu)]^{-1}$
(see Section~\ref{sec:regression} for details). Figure~\ref{fig:families} compares the
cumulant and mean functions for Bernoulli/logistic regression, SVM, and
Soft-SVM with $\kappa = 5$ and $\delta=0.8$. Note how the Soft-SVM functions match the SVM
behavior for more extreme values of the canonical parameter $\theta$ and the
logistic regression behavior for moderate values of $\theta$, achieving a good
compromise as a function of the softness $\kappa$.

\subsection{Soft-SVM Regression}
\label{sec:regression}
Given our characterization of the Soft-SVM exponential family distribution,
performing Soft-SVM regression is equivalent to fitting a regularized
generalized linear model. To this end, we follow a cyclic gradient ascent
procedure, alternating between an iteratively re-weighted least squares (IRLS)
step for Fisher scoring~\cite{mccullagh89} to update the coefficients $\beta$
given the softness $\kappa$, and then updating $\kappa$ and $\alpha$ given the current
values of $\beta$ via a Newton-Raphson scoring step.

For the IRLS step we need the link $g$ and the variance function
$V$ in~\eqref{eq:softsvm-var}. The link is given by solving the composite function $f^{-1}_{\kappa,\delta}\circ[b^{'}_{\kappa,\delta}(\mu)]^{-1}$ for $\eta$, where
\[
[b^{'}_{\kappa,\delta}(\mu)]^{-1} =\frac{1}{\kappa}
\Bigg\{\frac{1}{2}\log\Big(\frac{\mu}{1 - \mu}\Big) +
\text{asinh}\big(e^{h_{\kappa,\delta}(\mu)}\big)\Bigg\},
\]
with
\[
h_{\kappa,\delta}(\mu) =\log\cosh(2\kappa \delta) +
\log\frac{|\mu - 1/2|}{\sqrt{\mu(1-\mu)}},
\]
and
\[
f^{-1}_{\kappa,\delta}(\theta) =\frac{\theta}{2} + \frac{1}{\kappa}
{\text{asinh}\big(e^{H_{\kappa,\delta}(\theta)}\big)} \quad\text{with}\quad
H_{\kappa,\delta}(\theta) = -\kappa\delta + \log\text{sinh}\Big(\frac{\kappa|\theta|}{2}\Big).
\]
Note that in all these expressions $\delta$ is always expressed in the product
$\kappa\delta$, so it is immediate to use the alternative parameter $\alpha$
instead of $\delta$.

Our algorithm is summarized in Algorithm~\ref{algo:softsvm}. For notation simplicity, we represent both bias and coefficients in a single vector $\beta$
and store all features in a design matrix $X$, including a column of ones for
the bias (intercept).
\begin{algorithm}
	\Input{Features in $n$-by-$(f + 1)$ design matrix $X$ (including intercept)
		and labels $\mathbf{y}$.}
	\Parameters{Penalty $\lambda$, convergence tolerance $\epsilon$, and small
		perturbation $0 < \nu < 1$.}
	\Output{Estimated regression coefficients $\hat{\beta}$ (including
		bias $\hat{\beta}_0$), shape parameter $\hat{\alpha}$ and softness parameter $\hat{\kappa}$.}
	\Comment{Initialization:}
	Set $\alpha^{(0)} = 1$, $\kappa^{(0)} = 1$ and, for $i = 1, \ldots, n$,
	$\mu_i^{(0)} = (y_i + \nu) / (1 + 2\nu)$ as a
	$\nu$-perturbed version of $\mathbf{y}$ and
	$\eta_i^{(0)} = g_{\kappa^{(0)}}(\mu_i^{(0)})$\;
	\Comment{Cyclic scoring iteration:}
	\For{$t = 0, 1, \ldots$ (until convergence)}{
		\Comment{$\kappa$-step:}
		Update $\kappa^{(t+1)} = \kappa^{(t)} -
		[\partial^2 \ell_{\kappa,\alpha}/\partial \kappa^2]^{-1}
		(\partial \ell_{\kappa,\alpha}/\partial \kappa)$ via Newton's method\;
        \Comment{$\alpha$-step:}
        Update $\alpha^{(t+1)} = \alpha^{(t)} -
        \sum_{i=1}^{N}[y_{i}f_{\kappa,\alpha}'-b_{\kappa,\alpha}']/\sum_{i=1}^{N}[y_{i}f_{\kappa,\alpha}''-b_{\kappa,\alpha}'']$ via Newton's method\;
		\Comment{$\beta$-step:}
		Compute weights $w_i = y_i f''_{\kappa,\alpha}(\eta_i) - f''_{\kappa,\alpha}(\eta_i) b'_{\kappa,\alpha}(\theta_{i}) - (f'_{\kappa,\alpha}(\eta_i))^{2}b''_{\kappa,\alpha}(\theta_{i})$ and
		set $W^{(t)} = \text{Diag}_{i=1,\ldots,n}\{w_i^{(t)}\}$\;
		Update $\beta^{(t+1)}$ by solving
		$\big(X^\top W^{(t)} X + \lambda (0 \oplus I_f)\big) \beta^{(t+1)} =
		X^\top W^{(t)} \eta^{(t)} - X^\top\text{Diag}_{i=1,\ldots,n}\{f_{\kappa,\alpha}^{'}(\eta_i)\}( \mathbf{y} - \bm{\mu}(\beta^{(t)}))$\;
		Set $\eta^{(t+1)} = X\beta^{(t+1)}$,
		$\bm{\mu}(\beta^{(t)}) = b^{'}_\kappa(\bm{\theta})$, and
		$\bm{\theta}= [\theta_{1}, \theta_{2}, \ldots, \theta_{n}]^\top$\;
		\If{$|\ell_{(\kappa,\alpha)^{(t+1)}}(\eta^{(t+1)}; \mathbf{y}) -
			\ell_{(\kappa,\alpha)^{(t)}}(\eta^{(t)}; \mathbf{y})| /
			|\ell_{(\kappa,\alpha)^{(t)}}(\eta^{(t)}; \mathbf{y})| < \epsilon$}{\textbf{break}}
	}
	\Return{$\beta^{(t)}$, $\alpha^{(t)}$ and $\kappa^{(t)}$.}
	\caption{Soft-SVM Regression}
	\label{algo:softsvm}
\end{algorithm}

\subsection{Soft-SVM Variance Function}
\label{sec:varfun}
Since $h_{\kappa, \delta}(\mu) = h_\alpha(\mu) = \log\cosh(2\alpha) + \log|\mu
- 1/2|/\sqrt{\mu(1-\mu)}$, we have that
\[
\kappa([b^{'}_{\kappa, \delta}(\mu)]^{-1} \pm 2\delta) =
\frac{1}{2}\log\Big(\frac{\mu}{1-\mu}\Big) + \text{asinh}\Big(e^{h_\alpha(\mu)}\Big)
\pm 2\alpha
\]
and so the variance function in~\eqref{eq:softsvm-var} can be written as
$V_{\kappa, \alpha}(\mu) = \kappa \cdot r_\alpha(\mu)$ where, with
$u_\alpha(\mu) \doteq \text{logit}(\mu)/2 +
\text{asinh}(\exp\{h_\alpha(\mu)\})$,
\[
r_\alpha(\mu) = \frac{1}{2}\Big\{
v(u_\alpha(\mu) + 2\alpha) + v(u_\alpha(\mu) - 2\alpha) \Big\}.
\]
This way, parameters $\kappa$ and $\alpha$ have specific effects on the
variance function: the softness parameter $\kappa$ acts as a dispersion,
capturing the scale of $V(\mu)$, while the scaled separation $\alpha$ controls
the shape of $V(\mu)$.
\begin{figure}
\centering
\includegraphics[width=1\textwidth]{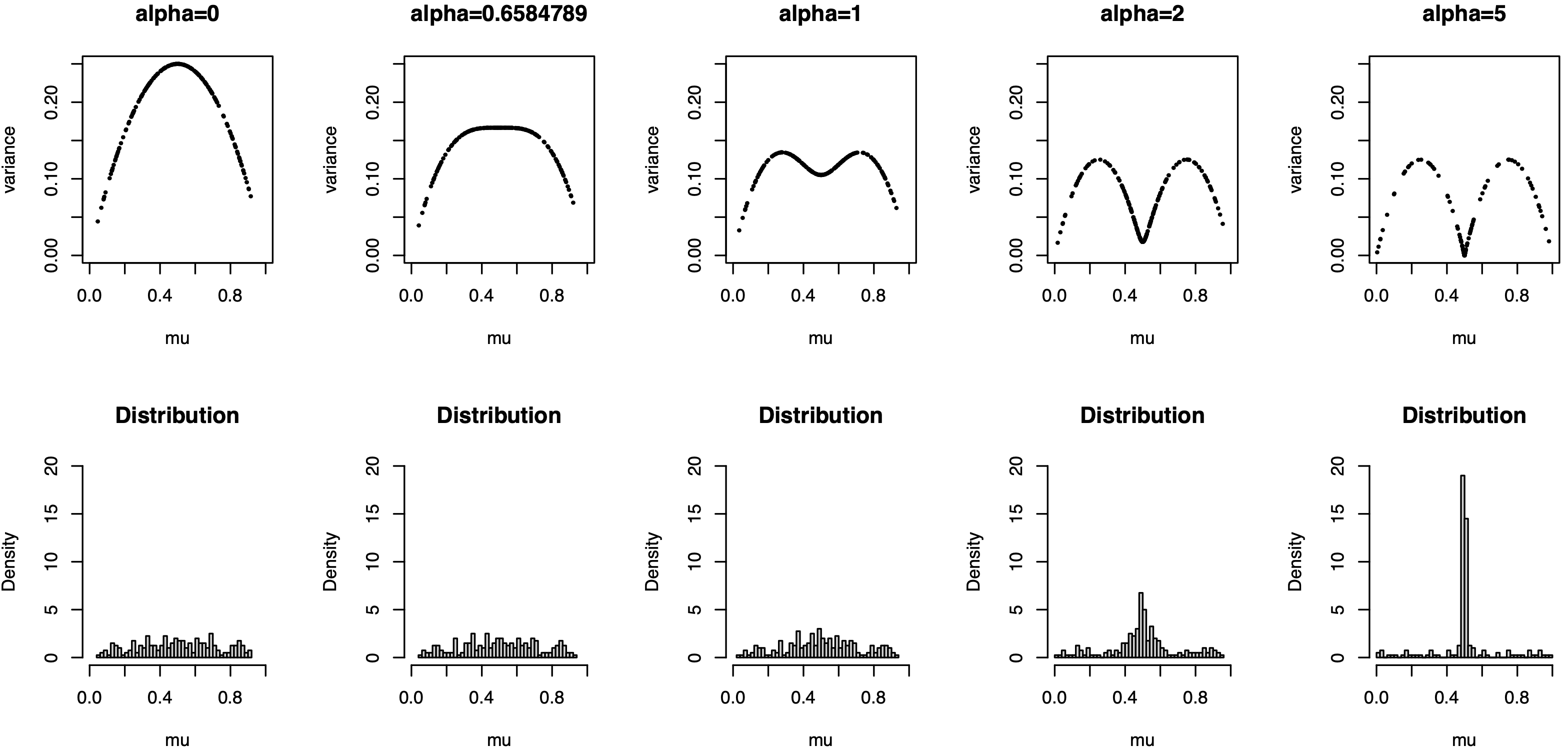}
\caption{Top row has variance-mean plots when scale parameter $\kappa$ is
fixed to $1$ and $\alpha \in \{0, \alpha^*, 1, 2, 5\}$, where $\alpha^*
\approx 0.66$ is the critical value when the peaks start separating. Bottom
row has histograms for the corresponding distribution of mean values.}
\label{fig:scale-shape}
\end{figure}

Figure~\ref{fig:scale-shape} shows variance function plots for the ESL mixture
dataset discussed in Section~\ref{sec:esl} when penalty parameter $\lambda$
and scale-parameter $\kappa$ are both fixed at $1$, and only allow the shape
parameter $\alpha$ to increase from $0$ to $5$. We can see that as $\alpha$
increases, the overall variance decreases and the plot gradually splits from a
unimodal shaped curve into a bimodal shaped curve. In practice, we observe
that as $\alpha$ increases to achieve higher mode separation, most
observations are pushed to have fitted $\mu$ values around $0.5$ and variance
weights close to zero, caught in-between variance peaks.

\begin{figure}
\centering
\includegraphics[width=1\textwidth]{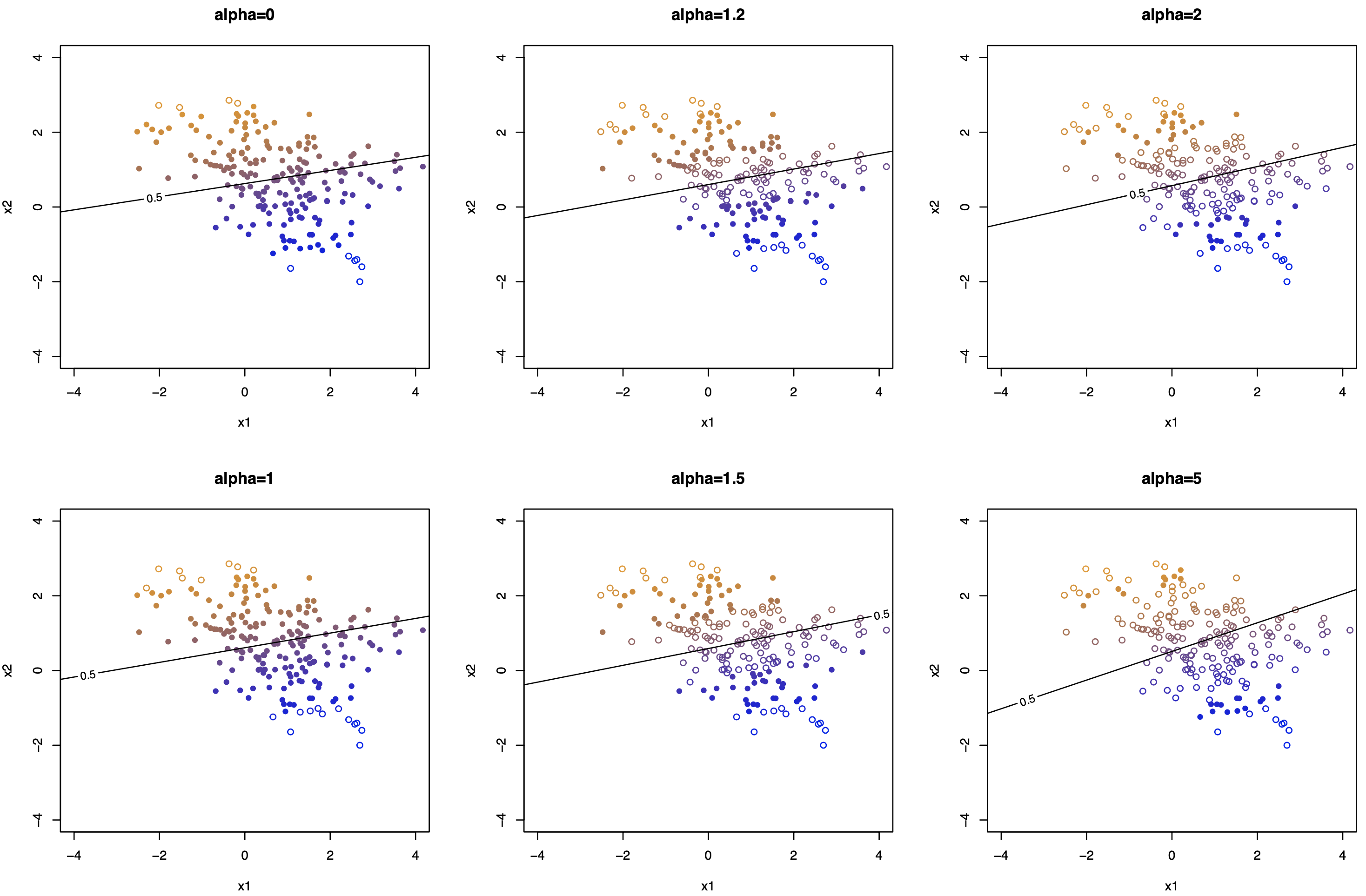}
\caption{Plots of classification results by using soft-SVM when scale
parameter $\kappa$ is fixed to 1 and $\alpha$ equals to 0, 1, 1.2, 1.5, 2,
and 5. The colors represent a gradient from blue ($\hat{\mu}_i = 0$) to
orange ($\hat{\mu}_i = 1)$. Filled dots have $V(\hat{\mu}_i) \geq 1$ while
hollow dots have $V(\hat{\mu}_i) < 1$.}
\label{fig:dead-zone}
\end{figure}

We can then classify observations with respect to their mean and variance
values: points with high variance weights have a large influence on the fit and
thus on the decision boundary, behaving like ``soft'' support vectors; points
with low variance weights and mean close to $0.5$ are close to the boundary
but do not influence it, belonging thus to a ``dead zone'' for the fit;
finally, points with low variance weights and large $\max\{\mu, 1-\mu\}$ are
inliers and have also low influence in the fit. This classification is
illustrated in Figure~\ref{fig:dead-zone}.


\subsection{Handling numerical issues}
As $\kappa$ increases we should anticipate numerical issues as $b_\kappa$ and $f_\kappa$ becomes non-differentiable up to machine precision. For this
reason, we need to adopt numerically stable versions for $\text{asinh}(e^x)$, $\log \cosh(x)$ and $\log\sinh(x)$, needed by $g_\kappa$ and thus also by $V_\kappa$
above.

The first step is a numerically stable version for the Bernoulli cumulant;
with $\epsilon$ the machine precision, we adopt
\[
\text{log1pe}(x) \doteq \log(1+e^{x}) = \left\{
\begin{array}{rcl}
x, & \mbox{if} & x > -\log\epsilon \\
x + \log(1+e^{-x}), & \mbox{if} & 0 < x \le -\log\epsilon \\
\log(1+e^{x}), & \mbox{if} & \log\epsilon \le x \le 0\\
0, &\mbox{if} & x < \log\epsilon.
\end{array}\right.
\]
This way, the soft plus can be reliably computed as
$p_\kappa(x) = \text{log1pe}(\kappa x) / \kappa$. Again exploiting domain
partitions and the parity of cosh, we can define
\[
\log\cosh(x) = \left\{
\begin{array}{rcl}
|x| - \log 2, & \mbox{if} & |x| > -\log\sqrt{\epsilon} \\
|x| - \log 2 + \text{log1pe}(-2x), & \mbox{if} & |x| \leq -\log\sqrt{\epsilon}. \\
\end{array}\right.
\]
We branch twice for $\text{asinh}(e^x)$: we first define
\[
\gamma(x) = \left\{
\begin{array}{rcl}
1, & \mbox{if} & |x| > -\log\sqrt{\epsilon} \\
\sqrt{1 + e^{-2|x|}}, & \mbox{if} & |x| \leq -\log\sqrt{\epsilon}, \\
\end{array}\right.
\]
and then
\[
\text{asinh}(e^x) = \left\{
\begin{array}{rcl}
x + \log(1 + \gamma(x)), & \mbox{if} & x > 0 \\
\log(e^x + \gamma(x)), & \mbox{if} & x \leq 0. \\
\end{array}\right.
\]
Finally, we write
\[
\log\sinh(x)= \left\{
\begin{array}{rcl}
\log\frac{1}{2} + x + \log(1-e^{-2x}), & \mbox{if} &x >  0\\
\log\frac{1}{2} - x + \log(1-e^{2x}), & \mbox{if} &x \le 0\\
\end{array}\right.
\]
to make it stable whenever $x$ is greater or smaller than $0$.

\subsection{Practical example: ESL mixture}
\label{sec:esl}
\begin{figure}
\centering
\includegraphics[width=0.9\textwidth]{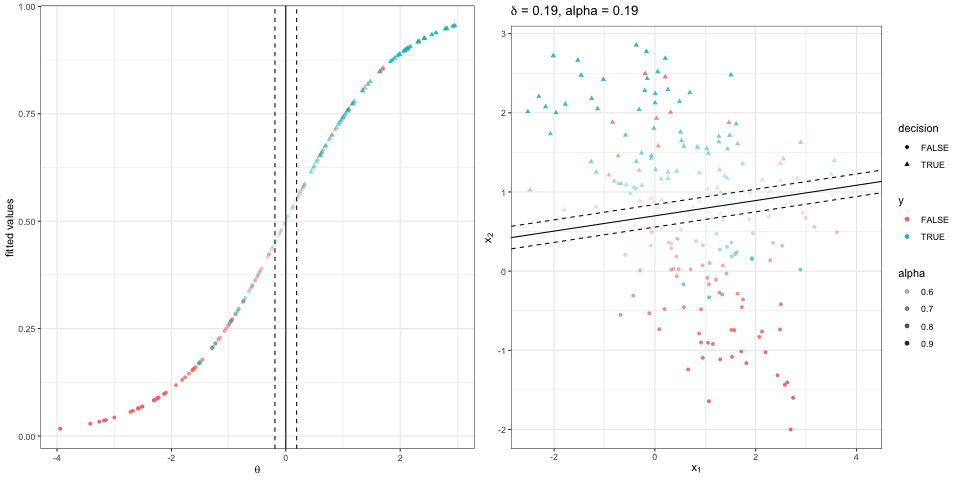}
\caption{Soft-SVM regression on ESL mixture data: fitted values $\hat{y}$
(left) and decision boundary in feature space (right). Points are shaped by predicted
classification, colored by observed labels, and have transparency proportional
to $|\hat{y}_i - 0.5|$. Hashed lines mark $\delta$ in the fitted
value plot and the soft margin $M = \delta / \|\hat{\beta}\|_2$
in the feature space plot. $\delta=\alpha/\kappa$.}
\label{fig:esl-example}
\end{figure}

To exam how the Soft-SVM regression works for classification in
comparison with SVM and logistic regression, we carry out a detailed example
using the simulated mixture data in~\cite{hastie09}, consisting of two
features, $X_1$ and $X_2$. To estimate the complexity penalty $\lambda$ we use
ten-fold cross-validation with 20 replications for all methods. The results
are summarized in Figure~\ref{fig:esl-example}. Because $\hat{\kappa}$ is not
much larger than unit, the fitted curve resembles a logistic curve; however,
the softness effect can be seen in the right panel as points with the soft
margin have estimated probabilities close to $0.5$, with fitted values quickly
increasing as points are farther away from the decision boundary. To classify,
we simply take a consensus rule, $\hat{y}_i = I(\hat{\mu}_i > 0.5)$, with $I$
being the indicator function.

\begin{figure}
\centering
\includegraphics[width=0.9\textwidth]{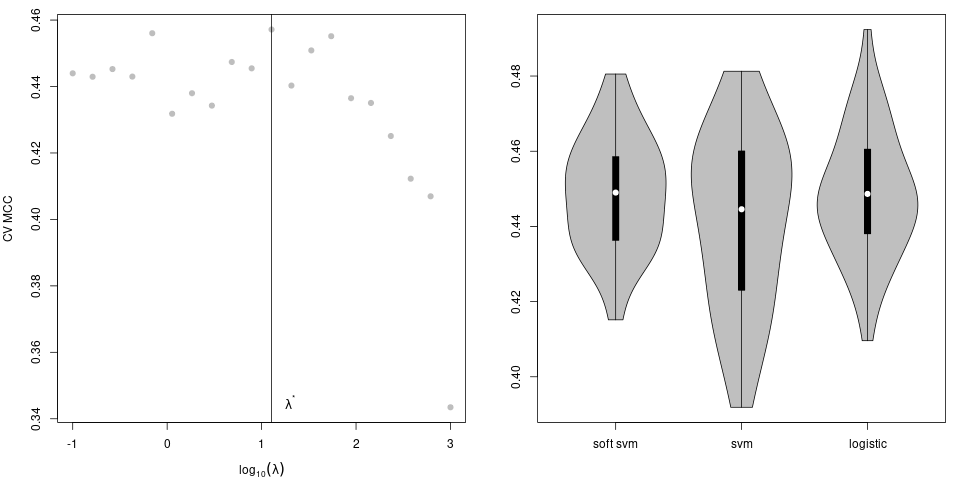}
\caption{Cross-validation results (replication averages) for Soft-SVM using
MCC (left) and performance comparisons to SVM and logistic regression
(right).}
\label{fig:esl-example-comp}
\end{figure}

Figure~\ref{fig:esl-example-comp} shows the effect of regularization in the
cross-validation performance of Soft-SVM regression, as measured by Matthews
correlation coefficient (MCC). We adopt MCC as a metric because it achieves a
more balanced evaluation of all possible outcomes~\cite{chicco20}. As can be
seen in the right panel, comparing MCC with SVM and logistic regressions, the
performances are quite comparable; Soft-SVM regression seems to have a slight
advantage on average, however, and logistic regression has a more variable
performance.

\section{Experiments}
\label{sec:experiments}

\subsection{Simulation study}
Prior works have shown that the performance of SVM can be severely affected
when applied to unbalanced datasets since the margin obtained is biased
towards the minority class~\cite{bernau14,cervantes20}. In this section, we
conduct a simulation study to investigate these effects on Soft-SVM
regression.

We consider a simple data simulation setup with two features. There are $n$
observations, partitioned into two sets with sizes $n_1 = \lfloor \rho n
\rfloor$ and $n_2 = n - n_1$, for an imbalance parameter $0 < \rho \leq 0.5$.
We expect to see SVM performing worse as $\rho$ gets smaller and the sets
become more unbalanced. Observations in the first set have labels $y_i = 0$
and $x_i \sim N(\mu_1, \sigma I_2)$, while the $n_2$ observations in the
second set have $y_i = 1$ and $x_i \sim N(\mu_2, \sigma I_2)$. We set
$\mu_1 = (\sqrt{2}, 1)$ and $\mu_2 = (0, 1 + \sqrt{2})$, so that the
decision boundary is $X_2 = X_1 + 1$ and the distance from either $\mu_1$ or
$\mu_2$ to the boundary is $1$. Parameter $\sigma$ acts as a measure of
separability: for $\sigma < 1$ we should have a stronger performance from the
classification methods, but as $\sigma$ becomes too small we expect to see
complete separability challenging logistic regression.

\begin{figure}
\centering
\includegraphics[width=0.9\textwidth]{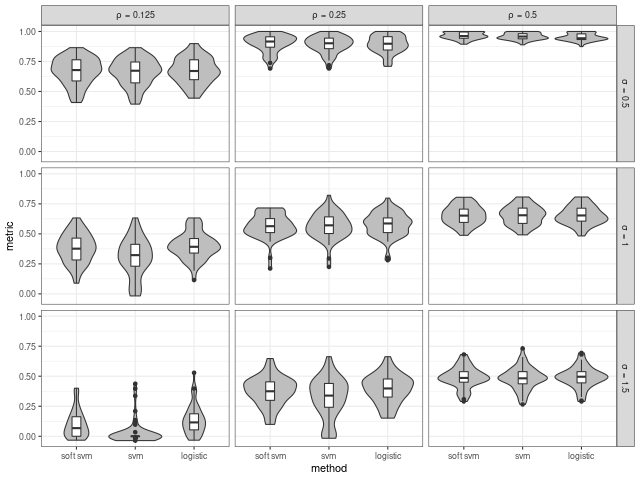}
\caption{Simulation study results comparing MCC for Soft-SVM, SVM, and
logistic regressions for $\rho \in \{0.125, 0.25, 0.5\}$ and
$\sigma \in \{0.5, 1, 1.5\}$.}
\label{fig:sim-example}
\end{figure}

For the simulation study we take $n = 100$ and simulate $50$ replications
using a factorial design with $\rho \in \{0.12, 0.25, 0.5\}$ and
$\sigma \in \{0.5, 1, 1.5\}$. For the SVM and Soft-SVM classification methods
we estimate complexity penalties $\lambda$ using ten-fold cross-validation and
use standard, non-regularized logistic regression. We report performance using
MCC and summarize the results in Figure~\ref{fig:sim-example}.

The performance across methods is comparable, uniformly degrading as we
boost confounding by decreasing $\rho$ and increasing $\sigma$, but differs in
more pathological corner cases. The combination of small $\rho$ values (higher
imbalance) and large $\sigma$ values (less separability) results in a poorer
performance from SVM, as previously reported~\cite{cervantes20}. Small
$\sigma$ values, on the other hand, result in lack of convergence and inflated
coefficient estimates in logistic regression fits, even though predicted
classification is not much affected. Soft-SVM classification, however, avoids
both issues, exhibiting stable estimation and good performance in all cases.

\subsection{Case studies}
To evaluate the empirical performance of Soft-SVM regression and
classification, we select nine well-known datasets from the UCI machine
learning repository~\cite{dua19} with a varied number of observations and
features. Table~\ref{tab:cases} lists the datasets and their attributes. Not
all datasets are originally meant for binary classification; dataset Abalone,
for instance, has three classes---male, female and infant---so we only take
male and female observations and exclude infants. The red/white wine quality
datasets originally have ten quality levels, labeled from low to high. We
dichotomize these levels into a low quality class (quality levels between $1$
and $5$, inclusive) and a high quality class (quality levels $6$ to $10$). All
remaining datasets have two classes.

\begin{table}
\begin{center}
\caption{Selected case study datasets from UCI repository.}
\label{tab:cases}
\begin{tabular}{lcc}
\toprule
\textbf{UCI dataset} & \textbf{\#~of observations} $n$ & \textbf{\#~of features} $f$ \\
\midrule
Abalone (only F \&\ M classes) & 4177 & 8 \\
Australian credit & 690 & 14 \\
Breast cancer & 569 & 30 \\
Haberman & 306 & 3 \\
Heart disease & 303 & 13 \\
Liver disorder & 345 & 6 \\
Pima Indians diabetes & 768 & 8 \\
Red wine quality (amended) & 1599 & 11 \\
White wine quality (amended) & 4898 & 11 \\
\bottomrule
\end{tabular}
\end{center}
\end{table}

\begin{figure}
\includegraphics[width=0.9\textwidth]{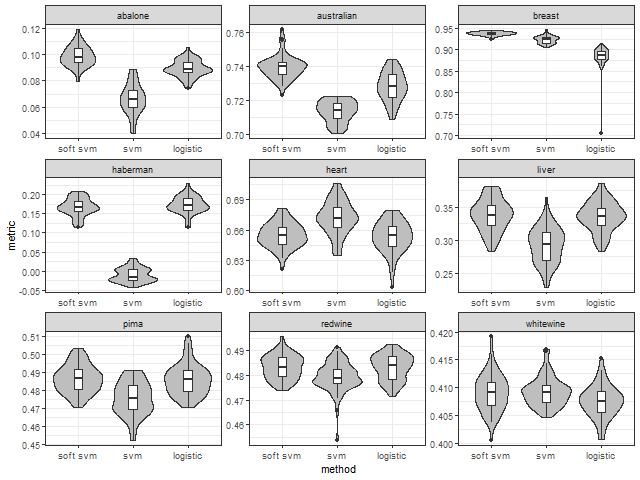}
\centering
\caption{Performance comparison of MCC for Soft-SVM, SVM, and logistic
regressions for nine UCI datasets over $50$ ten-fold cross validation studies.}
\label{fig:cases}
\end{figure}

As in the simulation study, we assess performance using MCC from $50$
ten-fold cross-validation replications. For each replication, we estimate
penalties $\lambda$ for all classification methods again using ten-fold
cross-validations. Figure~\ref{fig:cases} shows the results.

In general, it can be seen that SVM classification performs worse on ``tough''
datasets with low MCC performance, such as Abalone and Haberman, probably due to poor selection of penalty parameters. On the other hand, SVM performs
favorably when compared to Soft-SVM and logistic regressions in the Heart
dataset, being more robust to cases where many features seem to be
non-informative of classification labels. Overall, Soft-SVM regression
performs comparably to the best of SVM and logistic regression even in hard
datasets, as it was observed in the simulation study. Interestingly, it has a
clear advantage over both SVM and logistic regressions in cases where a few
features are very informative, such as in the Abalone and Australian credit
datasets. It often has an edge over logistic regression in over-dispersed
datasets such as Breast cancer and White wine datasets. It seems that Soft-SVM regression can better accommodate these cases via higher dispersion on moderate values of softness, i.e. $\kappa < 2$, or via better use of
regularization for increased robustness and generalizability.
\section{Conclusion}
We have presented a new binary classification and regression method based on
a new exponential family distribution, Soft-SVM. Soft-SVM arises from
extending a convex relaxation of the SVM hinge loss with a softness parameter,
in effect achieving a smooth transformation between the binomial deviance and
SVM hinge losses. Soft-SVM regression then entails fitting a generalized
linear model, and so it is comparable to logistic regression in terms of
computational complexity, having just a small overhead to fit the softness
parameter. Moreover, since many points might have low variance weights by not
being ``soft'' support vectors---either being dead-zoners or inliers, by our
classification in Section~\ref{sec:varfun}---they can be selected out of the
fit to achieve significant computational savings.

More importantly, as we have shown in a simulation study and multiple case
studies, Soft-SVM classification performs comparably to the best of SVM and
logistic regressions, often over performing them, even if slightly, in
feature-rich datasets. This advantage in performance can be attributed to
higher robustness from the softness and separation parameters, enabling the model to
accommodate well over-dispersed datasets and adequately address separability
issues via regularization. By bridging the SVM hinge and binomial deviance
loss functions, it seems that Soft-SVM regression is able to combine
robustness and interpretability, the advantages of SVM and logistic
regression, respectively, freeing the practitioner from the burden of
selecting one or the other.

For future work, we intend to develop data sampling procedures and adopt a
Bayesian formulation to then sample posterior estimates of coefficients,
softness and separation parameters. Another natural direction of work is to incorporate
kernels, as it is common in SVM regression. Finally, we also expect to extend
this formulation to multinomial classification~\cite{zhu05}, hoping again to
bring some robustness from SVM while keeping interpretability.

\bibliographystyle{plainnat}
\bibliography{neurips_2021}

\newpage

\appendix
\section{Appendix}
\subsection{\boldmath$[b^{'}_{\kappa}(\mu)]^{-1}$}
%
\begin{align*}
[b^{'}_{\kappa}(\mu)]^{-1} &= \frac{1}{\kappa} \log\Bigg\{\frac{-(2\mu-1)(e^{-2\kappa\delta(\kappa)} + e^{2\kappa\delta(\kappa)})}{2(2\mu-2)} + \sqrt{[\frac{-(2\mu-1)(e^{-2\kappa\delta(\kappa)} + e^{2\kappa\delta(\kappa)})}{2(2\mu-2)}]^{2} - \frac{\mu}{1-\mu}}  \Bigg\}\\
&=\frac{1}{\kappa} \log\Bigg\{\sqrt{\frac{\mu}{1 - \mu}} (\frac{(\mu-\frac{1}{2})\frac{1}{2}(e^{-2\kappa\delta(\kappa)} + e^{2\kappa\delta(\kappa)}) }{\sqrt{\mu(1-\mu)}}+ \sqrt{[\frac{(\mu-\frac{1}{2})\frac{1}{2}(e^{-2\kappa\delta(\kappa)} + e^{2\kappa\delta(\kappa)}) }{\sqrt{\mu(1-\mu)}}]^{2} + 1} ) \Bigg\}\\
&=\left\{
\begin{array}{rcl}
\frac{1}{\kappa} [\frac{1}{2}\log\frac{\mu}{1 - \mu} + \log(e^{h_\kappa(\mu)} + \sqrt{e^{2h_\kappa(\mu)} + 1}) ], & \mbox{if} &\mu >  \frac{1}{2}\\
\frac{1}{\kappa} [\frac{1}{2}\log\frac{\mu}{1 - \mu} + \log(-e^{h_\kappa(\mu)} + \sqrt{e^{2h_\kappa(\mu)} + 1}) ], & \mbox{if} &\mu\le \frac{1}{2}
\end{array}\right.\\
&=\frac{1}{\kappa} \Bigg\{\frac{1}{2}\log(\frac{\mu}{1 - \mu}) +
\text{asinh}\big(e^{h_\kappa(\mu)}\big)\Bigg\},
\end{align*}
\[
h_\kappa(\mu) = \log\Bigg\{ \frac{|\mu-\frac{1}{2}|}{\sqrt{\mu(1-\mu)}}\cdot \frac{1}{2}(e^{-2\kappa\delta(\kappa)} + e^{2\kappa\delta(\kappa)})\Bigg\}
=\log\cosh(2\kappa \delta(\kappa)) +
\log\frac{|\mu - 1/2|}{\sqrt{\mu(1-\mu)}}.
\]

\subsection{\boldmath$f^{-1}_{\kappa}(\theta)$}
\begin{align*}
f^{-1}_{\kappa}(\theta) &= \frac{1}{\kappa} \log\Bigg\{\frac{e^{\kappa\theta}}{2e^{\kappa\delta(\kappa)}} + \sqrt{e^{\kappa\theta} + (\frac{e^{\kappa\theta} -1}{2e^{\kappa\delta(\kappa)}})^2}  \Bigg\}\\
&= \frac{1}{\kappa} \log\Bigg\{(e^{\kappa\theta})^{\frac{1}{2}}\frac{e^{\kappa\theta}-1}{2e^{\kappa\delta(\kappa)}(e^{\kappa\theta})^{\frac{1}{2}}} + (e^{\kappa\theta})^{\frac{1}{2}}\sqrt{1 + (\frac{e^{\kappa\theta} -1}{2e^{\kappa\delta(\kappa)}(e^{\kappa\theta})^{\frac{1}{2}}})^2}  \Bigg\}\\
&=\frac{1}{\kappa} \log\Bigg\{(e^{\kappa\theta})^{\frac{1}{2}}\Bigg\} +
\frac{1}{\kappa}\log\Bigg\{\frac{e^{\kappa\theta}-1}{2e^{\kappa\delta(\kappa)}(e^{\kappa\theta})^{\frac{1}{2}}} + \sqrt{1 + (\frac{e^{\kappa\theta} -1}{2e^{\kappa\delta(\kappa)}(e^{\kappa\theta})^{\frac{1}{2}}})^2}  \Bigg\}\\
&= \left\{
\begin{array}{rcl}
\frac{\theta}{2} + \frac{1}{\kappa}\log(e^{H_\kappa(\theta)} + \sqrt{e^{2H_\kappa(\theta)} + 1}) ], & \mbox{if} &\theta >  0 \\
\frac{\theta}{2} + \frac{1}{\kappa}\log(-e^{H_\kappa(\theta)} + \sqrt{e^{2H_\kappa(\theta)} + 1}) ], & \mbox{if} &\theta \le  0\\
\end{array}\right.\\
&=\frac{\theta}{2} + \frac{1}{\kappa} {\text{asinh}\big(e^{H_\kappa(\theta)}\big)},
\end{align*}
\begin{align*}
H_\kappa(\theta) &= \log\Bigg\{ e^{-\kappa\delta(\kappa)}\cdot\frac{|e^{\kappa\theta}-1|}{2(e^{\kappa\theta})^{\frac{1}{2}}}\Bigg\}\\
&=-\kappa\delta(\kappa) + \log\text{sinh}(\frac{\kappa|\theta|}{2}),
\end{align*}
\[
\log\text{sinh(x)}=\log\Bigg(\frac{e^{2x}-1}{2e^{x}}\Bigg)= \left\{
\begin{array}{rcl}
\log\frac{1}{2} + x + \log(1-e^{-2x}), & \mbox{if} &x >  0\\
\log\frac{1}{2} - x + \log(1-e^{2x}), & \mbox{if} &x \le 0\\
\end{array}\right..
\]

\subsection{\boldmath$V_\kappa(\mu)$}
\[
V_\kappa(\mu) = b_\kappa''(\theta) = b_\kappa''\big(g_\kappa(\mu)\big),
\]
\begin{align*}
b_\kappa''(\theta) &= \frac{\partial^2\Biggl\{ 0.5*\Bigg[\frac{\log(1+e^{\kappa(\theta -2\delta(\kappa))})}{\kappa} +  \frac{\log(1+e^{\kappa(\theta +2\delta(\kappa))})}{\kappa}  \Bigg] \Biggr\} }{\partial^2 \theta}\\
&=\frac{\partial\Biggl\{  0.5*\Bigg[\frac{e^{\kappa(\theta - 2\delta(\kappa))}}{1 + e^{\kappa(\theta - 2\delta(\kappa))}}   + \frac{e^{\kappa(\theta + 2\delta(\kappa))}}{1 + e^{\kappa(\theta + 2\delta(\kappa))}}   \Bigg] \Biggr\}}{\partial \theta}\\
&=0.5*\Bigg[\frac{\kappa e^{\kappa(\theta - 2\delta(\kappa))}}{(1 + e^{\kappa(\theta - 2\delta(\kappa))})^2}   + \frac{\kappa e^{\kappa(\theta + 2\delta(\kappa))}}{(1 + e^{\kappa(\theta + 2\delta(\kappa))})^2}   \Bigg]\\
&=\frac{\kappa}{2}\biggl\{v[\kappa(\theta -2\delta(\kappa))] + v[\kappa(\theta +2\delta(\kappa))]  \biggr\},
\end{align*}
\[
\delta(\kappa)=1-\frac{1}{\kappa},\  v(x)=\frac{e^{x}}{1+e^{x}}.
\]

\subsection{\boldmath$\beta$-step}
\[
l=\sum_{i=1}^{n}y_i\theta_{i} - b_{\kappa}(\theta_{i}),\ \theta_{i}=f_{\kappa}(\eta_i), \ y_i=X_i^\top\bm{\beta}= \beta_0+ x_i\beta_1+\ldots+x_p\beta_p,], \]
\begin{align*}
U(\beta_j)_{j=0,1,\ldots,p} &= \frac{\partial l}{\partial \beta_j}= \sum_{i=1}^{n}\frac{\partial(y_i\theta_{i}- b_{\kappa}(\theta_{i})))}{\partial\theta_{i}} \cdot \frac{\partial\theta_{i}}{\beta_j}\\
&=\sum_{i=1}^{n}(y_i- \bm{\frac{\partial b_{\kappa}(\theta_{i})}{\partial \theta_{i}} \doteq b'_{\kappa}(\theta_{i})})\cdot f'_{\kappa}(y_i) \cdot X_{ij}\\
&=\sum_{i=1}^{n} f'_{\kappa}(y_i) \cdot X_{ij} \cdot (y_i - b'_{\kappa}(\theta_{i})),
\end{align*}
\begin{align*}
U(\bm{\beta})&=\frac{\partial l}{\partial \bm{\beta}}=  \begin{bmatrix}
U(\beta_0)\\
U(\beta_1)\\
...\\
U(\beta_p)\\
\end{bmatrix}
= \begin{bmatrix}
\sum_{i=1}^{n} f'_{\kappa}(y_i) \cdot X_{i0} \cdot (y_i - b'_{\kappa}(\theta_{i}))\\
\sum_{i=1}^{n} f'_{\kappa}(y_i) \cdot X_{i1} \cdot (y_i - b'_{\kappa}(\theta_{i}))\\
...\\
\sum_{i=1}^{n} f'_{\kappa}(y_i) \cdot X_{ip} \cdot (y_i - b'_{\kappa}(\theta_{i}))\\
\end{bmatrix}
=X^\top \text{Diag}_{i=1,\ldots,n}\{f_\kappa^{'}(\eta_i)\}\begin{bmatrix}
y_1 - b'_{\kappa}(\theta_{1})\\
y_2 - b'_{\kappa}(\theta_{2})\\
...\\
y_n - b'_{\kappa}(\theta_{n})\\
\end{bmatrix}\\
&=X^\top \text{Diag}_{i=1,\ldots,n}\{f_\kappa^{'}(\eta_i)\}(\bm{y - \mu(\theta)}), \text{where}\ \bm{\mu(\theta)} = \begin{bmatrix}
b'_{\kappa}(\theta_{1})\\
b'_{\kappa}(\theta_{2})\\
...\\
b'_{\kappa}(\theta_{n})\\
\end{bmatrix}.\\
\end{align*}
\begin{align*}
H(\bm{\beta})=\frac{\partial U(\bm{\beta})}{\partial \bm{\beta^\top}}= \frac{\partial \begin{bmatrix}
\sum_{i=1}^{n}	f'_{\kappa}(y_i) \cdot X_{i0} \cdot (y_i - b'_{\kappa}(\theta_{i}))\\
\sum_{i=1}^{n}	f'_{\kappa}(y_i) \cdot X_{i1} \cdot (y_i - b'_{\kappa}(\theta_{i}))\\
...\\
\sum_{i=1}^{n}	f'_{\kappa}(y_i) \cdot X_{ip} \cdot (y_i - b'_{\kappa}(\theta_{i}))\\
\end{bmatrix}}{\partial \bm{\beta^\top}}.
\end{align*}
For each row $j, j=0,1,2,...,p$:
\begin{align*}
\frac{\partial \sum_{i=1}^{n}	f'_{\kappa}(y_i) \cdot X_{ij} \cdot (y_i - b'_{\kappa}(\theta_{i}))}{\partial \bm{\beta^\top}} &=\frac{\partial \sum_{i=1}^{n}f'_{\kappa}(\eta_i)X_{ij}y_i}{\partial \bm{\beta^\top}} + \frac{\partial \sum_{i=1}^{n}f'_{\kappa}(\eta_i)X_{ij} b'_{\kappa}(\theta_{i})}{\partial \bm{\beta^\top}} \\
&=\begin{bmatrix}
\sum_{i=1}^{n} X_{ij}[y_i f''_{\kappa}(\eta_i) - f''_{\kappa}(\eta_i) b'_{\kappa}(\theta_{i}) - (f'_{\kappa}(\eta_i))^{2}b''_{\kappa}(\theta_{i})]X_{i0}\\
\sum_{i=1}^{n} X_{ij}[y_i f''_{\kappa}(\eta_i) - f''_{\kappa}(\eta_i) b'_{\kappa}(\theta_{i}) - (f'_{\kappa}(\eta_i))^{2}b''_{\kappa}(\theta_{i})]X_{i1}\\
...\\
\sum_{i=1}^{n} X_{ij}[y_i f''_{\kappa}(\eta_i) - f''_{\kappa}(\eta_i) b'_{\kappa}(\theta_{i}) - (f'_{\kappa}(\eta_i))^{2}b''_{\kappa}(\theta_{i})]X_{ip}\\
\end{bmatrix}^\top.\\
\end{align*}
Therefore,
\begin{align*}
H(\bm{\beta})&=\begin{bmatrix}
\begin{bmatrix}
\sum_{i=1}^{n} X_{i0}[y_i f''_{\kappa}(\eta_i) - f''_{\kappa}(\eta_i) b'_{\kappa}(\theta_{i}) - (f'_{\kappa}(\eta_i))^{2}b''_{\kappa}(\theta_{i})]X_{i0}\\
\sum_{i=1}^{n} X_{i0}[y_i f''_{\kappa}(\eta_i) - f''_{\kappa}(\eta_i) b'_{\kappa}(\theta_{i}) - (f'_{\kappa}(\eta_i))^{2}b''_{\kappa}(\theta_{i})]X_{i1}\\
...\\
\sum_{i=1}^{n} X_{i0}[y_i f''_{\kappa}(\eta_i) - f''_{\kappa}(\eta_i) b'_{\kappa}(\theta_{i}) - (f'_{\kappa}(\eta_i))^{2}b''_{\kappa}(\theta_{i})]X_{ip}\\
\end{bmatrix}^\top\\
\begin{bmatrix}
\sum_{i=1}^{n} X_{i1}[y_i f''_{\kappa}(\eta_i) - f''_{\kappa}(\eta_i) b'_{\kappa}(\theta_{i}) - (f'_{\kappa}(\eta_i))^{2}b''_{\kappa}(\theta_{i})]X_{i0}\\
\sum_{i=1}^{n} X_{i1}[y_i f''_{\kappa}(\eta_i) - f''_{\kappa}(\eta_i) b'_{\kappa}(\theta_{i}) - (f'_{\kappa}(\eta_i))^{2}b''_{\kappa}(\theta_{i})]X_{i1}\\
...\\
\sum_{i=1}^{n} X_{i1}[y_i f''_{\kappa}(\eta_i) - f''_{\kappa}(\eta_i) b'_{\kappa}(\theta_{i}) - (f'_{\kappa}(\eta_i))^{2}b''_{\kappa}(\theta_{i})]X_{ip}\\
\end{bmatrix}^\top\\
\vdots\\
\begin{bmatrix}
\sum_{i=1}^{n} X_{ip}[y_i f''_{\kappa}(\eta_i) - f''_{\kappa}(\eta_i) b'_{\kappa}(\theta_{i}) - (f'_{\kappa}(\eta_i))^{2}b''_{\kappa}(\theta_{i})]X_{i0}\\
\sum_{i=1}^{n} X_{ip}[y_i f''_{\kappa}(\eta_i) - f''_{\kappa}(\eta_i) b'_{\kappa}(\theta_{i}) - (f'_{\kappa}(\eta_i))^{2}b''_{\kappa}(\theta_{i})]X_{i1}\\
...\\
\sum_{i=1}^{n} X_{ip}[y_i f''_{\kappa}(\eta_i) - f''_{\kappa}(\eta_i) b'_{\kappa}(\theta_{i}) - (f'_{\kappa}(\eta_i))^{2}b''_{\kappa}(\theta_{i})]X_{ip}\\
\end{bmatrix}^\top\\
\end{bmatrix}\\
&=X^\top \text{Diag}_{i=1,\ldots,n}\{ \bm{y_i f''_{\kappa}(\eta_i) - f''_{\kappa}(\eta_i) b'_{\kappa}(\theta_{i}) - (f'_{\kappa}(\eta_i))^{2}b''_{\kappa}(\theta_{i})\doteq w_i} \}X\\
&=X^\top (\bm{\text{Diag}_{i=1,\ldots,n}\{w_i \}\doteq W(\beta)})X\\
&=X^\top W(\beta)X.
\end{align*}
Since by Newton's method we have:
\[
\beta^{(t+1)} \approx \beta^{(t)} - [H^{(t)}]^{-1} U(\beta^{(t)})
\Rightarrow H^{(t)} \beta^{(t+1)} \approx H^{(t)} \beta^{(t)} -  U(\beta^{(t)},
\]
and by adding the penalty term and plugging in the value of $H$ and $U$, we can update $\beta^{(t+1)}$ by solving the following matrix equation:
\[
\big(X^\top W^{(t)} X + \lambda (0 \oplus I_f)\big) \beta^{(t+1)} =
X^\top W^{(t)} \eta^{(t)} - X^\top\text{Diag}_{i=1,\ldots,n}\{f_\kappa^{'}(\eta_i)\}( \mathbf{y} - \bm{\mu}(\beta^{(t)})),\ \eta^{(t)} = X\beta^{(t)}.
\]

\end{document}